\documentclass{article}

\usepackage{PRIMEarxiv}

\usepackage[utf8]{inputenc} 
\usepackage[T1]{fontenc}    
\usepackage{hyperref}       
\usepackage{url}            
\usepackage{booktabs}       
\usepackage{amsfonts}       
\usepackage{nicefrac}       
\usepackage{microtype}      
\usepackage{lipsum}
\usepackage{fancyhdr}       
\usepackage{graphicx}       
\graphicspath{{media/}}     

\title{The Open-Source Advantage in Large Language Models (LLMs)}

\author{
  Jiya Manchanda \thanks{Equal contribution}\\
  Department of Philosophy and Religion \\
  Rollins College \\
  Winter Park\\
  \texttt{jmanchanda@rollins.edu} \\
  \and
  Laura Boettcher \thanks{Equal contribution}\\
  Department of Mathematics and Computer Science \\
  Rollins College \\
  Winter Park\\
  \texttt{lboettcher@rollins.edu} \\
  \and
  Matheus Westphalen \thanks{Equal contribution}\\
  Department of Mathematics and Computer Science \\
  Rollins College \\
  Winter Park\\
  \texttt{mwestphalen@rollins.edu} \\
  \and
  Jasser Jasser \\
  Department of Mathematics and Computer Science \\
  Rollins College \\
  Winter Park\\
  \texttt{jjasser@rollins.edu} \\
}

\begin{document}
\maketitle

\begin{abstract}

Large language models (LLMs) have rapidly advanced natural language processing, driving significant breakthroughs in tasks such as text generation, machine translation, and domain-specific reasoning. The field now faces a critical dilemma in its approach: closed-source models like GPT-4 deliver state-of-the-art performance but restrict reproducibility, accessibility, and external oversight, while open-source frameworks like LLaMA and Mixtral democratize access, foster collaboration, and support diverse applications, achieving competitive results through techniques like instruction tuning and LoRA. Hybrid approaches address challenges like bias mitigation and resource accessibility by combining the scalability of closed-source systems with the transparency and inclusivity of open-source framework. However, in this position paper, we argue that open-source remains the most robust path for advancing LLM research and ethical deployment.

\end{abstract}


\section{Introduction}
Large language models (LLMs) stand at the forefront of machine learning (ML) today, having revolutionized natural language processing (NLP) and driven advancements in tasks such as text generation, translation, domain-specific inferencing, and sentiment analysis \cite{minaee2024large}. These models have not only become indispensable tools in industry but also focal points of academic research. Early approaches to NLP were characterized by deterministic, rule-based systems that, while structurally explicit, were inherently rigid and limited in their adaptability \cite{di2016formal}. The introduction of probabilistic models, inspired by Shannon’s Information Theory, facilitated improved predictive performance by leveraging statistical correlations \cite{shannon1948mathematical}. However, these early statistical language models were limited in handling long-range dependencies and capturing the semantic properties of pragmatic language use \cite{bengio2000neural}.

A paradigmatic advancement in this domain was the development of the transformer architecture in 2017 by Vaswani et al., which introduced self-attention mechanisms to dynamically weight the significance of tokens within a sequence \cite{vaswani2017attention}. This breakthrough enabled large-scale scalability, overcoming the limitations of prior models like RNNs and LSTMs, which struggled with long-range dependencies and computational inefficiencies \cite{mienye2024recurrent}. The self-attention mechanism allowed for parallel sequence processing, drastically improving efficiency and expressiveness in NLP tasks \cite{jacobs2024system}. By training on extensive and diverse datasets, LLMs internalize sophisticated representations of language, allowing for the generation of contextually rich and syntactically coherent outputs. 

However, the rapid development of LLMs has brought into focus critical issues—chief among them is the tension between open-source and closed-source approaches \cite{yu2023open}. Closed-source LLMs are most commonly positioned as a means of safeguarding proprietary knowledge, ensuring security, and maintaining compliance with regulatory frameworks \cite{dong2024safeguarding}. Advocates of this approach contend that restricted access mitigates risks associated with adversarial manipulation, infringement upon intellectual property, and controlled dissemination of information \cite{kibriya2024privacy}. By contrast, open-source LLMs, by providing publicly accessible model architectures, training datasets, and algorithmic transparency, foster a collaborative research ecosystem that facilitates iterative development and rigorous external validation \cite{seger2023open}. Advocates of this approach emphasize the democratic expansion of machine learning (ML) capabilities, the mitigation of opaque biases through collective scrutiny, and the acceleration of advancements via diverse contributions \cite{shashidhar2023democratizing}. 

This underexplored division (in part, of labor), raises urgent normative questions as competing LLMs evolve in capability, accessibility, and governance. In particular, which approach to ML development the field should embrace to ensure both responsible innovation and sustained technological progress will be the subject of much consideration in this paper. In this position paper, we argue that open-source remains the most robust path for the advancement and ethical deployment of LLMs. Our argument relies on an evaluation of each approach based on four criteria: the rate and nature of innovations in development, the empirical performance benchmarks, reproducibility of results, and the degree of transparency in both methodology and implementation. We acknowledge that closed-source models are likely to remain prevalent; however, increasing pressure to adopt openness may lead to hybrid approaches that mediate both paradigms. Finally, we articulate directions for future research that highlight the unique opportunities offered in the development of open-source LLMs.

\section{Conceptual Foundations}
The conceptual landscape of large language model (LLM) development can be distinguished across three broad paradigms: closed-source, hybrid, and open-source. Each represents a philosophy of knowledge production and governance. Understanding these paradigms is thus essential to situating the ongoing debate over innovation, reproducibility, and accountability in artificial intelligence research.

Closed-source models represent the most traditional and restrictive approach. In this paradigm, both model weights and training corpora are held as proprietary assets, while key architectural details such as parameter counts, optimization strategies, and hyperparameter configurations are disclosed only partially, if at all. Interaction with the model is mediated through application programming interfaces (APIs) or controlled deployment environments, allowing users to benefit from capabilities without accessing internal mechanics. This arrangement is underpinned by a philosophy that regards LLMs as intellectual property. Development is organized around commitments to proprietary stewardship, control over distribution, and protection of commercially or strategically sensitive datasets. From this perspective, innovation in large-scale language models is a product of concentrated expertise and exclusive resource investment, and the resulting systems are maintained within boundaries that preserve institutional ownership and security.

Hybrid models mark a newer and increasingly significant middle ground. These systems adopt a framework of selective disclosure, making certain components of the model accessible while withholding others. For instance, weights may be released under specific licensing conditions, while pretraining corpora remain undisclosed; or fine-tuning protocols may be documented while full-scale optimization pipelines are retained as proprietary. The OpenAI gpt-oss series offers a clear example. The gpt-oss-120b is a 117-billion parameter model with 5.1 billion active parameters, designed for high-reasoning production tasks on a single 80GB GPU, distributed under controlled terms that limit modification and redistribution. By contrast, the gpt-oss-20b, a 21-billion parameter model with 3.6 billion active parameters, has been released fully open, enabling local deployment and domain-specific adaptation. The hybrid paradigm embodies a philosophy of conditional openness: a recognition of the value of accessibility and collaboration, paired with commitments to institutional stewardship and selective protection of resources.

Open-source models represent the most transparent approach. In this paradigm, model weights, architectural specifications, training logs, hyperparameters, and, where possible, training datasets are made openly available. These include Meta’s LLaMA-3 family, Mistral’s Mixtral, and the BigScience BLOOM model, each of which has been released with comprehensive documentation, checkpoints, and evaluation protocols. The guiding philosophy here is that LLMs are collective research artifacts and, to some extent, public goods. Open-source development rests on commitments to reproducibility, transparency, and the democratization of participation in machine learning research. By making models available for inspection, replication, and adaptation, this paradigm invites broad collaboration from academic, industrial, and independent researchers alike. It reflects an orientation toward modularity, inclusivity, and distributed innovation, where progress in architecture, fine-tuning, and evaluation is advanced through shared contributions.

\section{Innovations in Development}
The extent to which a given approach to LLM development—open-source or closed-source—facilitates, accelerates, and democratizes architectural and methodological breakthroughs is a critical metric for evaluating the trajectory of NLP research and deployment \cite{de2022neurolinguistic}. For, the ability to generate scalable, efficient, and adaptable models that enhance both accessibility and computational feasibility determines the sustainability and nature of progress we make.

Closed-source models have pioneered one of the most significant developments in recent years via scaling laws, as articulated by Kaplan et al. \cite{kaplan2020scaling}. These models demonstrated that increasing parameter counts, dataset breadth, and computational resources could yield non-linear performance improvements. However, empirical evidence suggests that unbounded scaling incurs diminishing returns beyond a certain threshold, necessitating intelligent resource allocation strategies \cite{chenscaling}. Consequently, research has shifted toward optimizing training paradigms such as curriculum learning and reinforcement learning from human feedback (RLHF) to improve efficiency without excessive computational expenditure \cite{lee2023rlaif}.

A significant architectural development in closed-source models is the Mixture-of-Experts (MoE) framework, which signifies a shift from dense to sparse computing \cite{puigcerver2023sparse}. Unlike traditional architectures that indiscriminately activate all model parameters, MoE partitions computational tasks among specialized subnetworks, or “experts,” selecting only a subset for any given input. This selective activation reduces computational redundancy while preserving expressive power. Google’s Switch Transformer exemplifies this approach, which was originally introduced in a research paper by Robert Jacobs and Geoffrey Hinton in 1991, illustrating that model capacity can be expanded without a proportional increase in computational cost \cite{jacobs1991adaptive}. However, closed-source MoE implementations generally employ static expert routing, where the assignment of inputs to experts follows predetermined heuristic patterns rather than adaptive optimization.

Closed-source models such as Open AI’s GPT and Anthropic’s Claude have refined transformer-based architectures through proprietary enhancements, including extensive pre-training on exclusive datasets \cite{gudibande2023false}. These refinements have led to significant improvements in fluency, coherence, and task adaptability. The release of GPT-3, with its 175 billion parameters, marked a milestone in scaling methodologies, demonstrating that large-scale architectures could enable few-shot learning across a broad array of NLP tasks \cite{floridi2020gpt}. However, despite these technical achievements, closed-source models remain highly opaque, restricting external scrutiny and equitable participation in AI development. This centralized control over computational and data resources has exacerbated global disparities in AI accessibility and innovation.

By contrast, the open-source approach has prioritized adaptive and efficiency-driven innovations, reducing reliance on brute-force scaling while promoting architectural modularity and computational accessibility. Models such as Meta’s LLaMA and BigScience’s BLOOM have demonstrated that strategic optimizations can rival or exceed the efficiency of closed-source alternatives without necessitating prohibitively large computational resources \cite{touvron2023llama}. A key advancement in open-source models has been dynamic expert routing in MoE architectures \cite{you2021speechmoe}. Mistral AI’s Mixtral 8x7B model has marked this shift by implementing a sparse MoE architecture where only 2 of 8 experts are activated per token, reducing computational overhead while maintaining performance \cite{wan2023efficient}. Likewise, DeepSeek and the Allen Institute’s Open Mixture-of-Experts Language Models (OLMoE) have developed adaptive MoE frameworks that allocate computational resources dynamically based on input complexity \cite{muennighoff2024olmoe}. This approach optimizes energy consumption while enhancing efficiency, making LLM deployment viable even for organizations with limited infrastructure. DeepSeek’s dynamic expert allocation further refines this concept by activating only the most relevant experts for a given input, thereby minimizing redundant computations and optimizing real-time execution \cite{liu2024deepseek}.

Moreover, open-source research has driven significant advancements in training optimizations, incorporating publicly available fine-tuning datasets such as Stanford’s Alpaca and Databricks’ Dolly to improve generalization across diverse applications \cite{chen2023alpagasus}, reducing the cost and hardware demands associated with fine-tuning LLMs \cite{shen2024rethinking}. Low-Rank Adaptation (LoRA) enables parameter-efficient tuning by freezing most model weights while optimizing a low-rank subset \cite{hu2021lora}, facilitating domain-specific adaptations such as ClinicalBERT for medical NLP and LEGAL-BERT for legal document analysis \cite{alsentzer2019publicly}. Another key development is that of knowledge distillation, a technique wherein a smaller “student” model is trained to replicate the performance of a larger “teacher” model, thereby compressing knowledge without sacrificing core functionalities \cite{gou2021knowledge}. DistilBERT exemplifies this strategy, offering a lightweight yet robust alternative to larger architectures. A third case of training optimization is the method of retrieval-augmented generation (RAG) \cite{lewis2020retrieval}. Unlike traditional language models that rely solely on pre-trained internal representations, RAG integrates external knowledge retrieval mechanisms, augmenting model responses with verifiable information. This approach helps mitigate against hallucinations and improves factual grounding, offering an alternative to parameter expansion for performance improvements \cite{niu2023ragtruth}. Open initiatives such as T5-RAG demonstrate how this development enables more scalable and knowledge-rich ML systems.

Beyond training methodologies, open-source models have contributed significantly to computational efficiency for the purpose of refining instruction-tuning and domain-specific adaptations. LLaMA’s grouped query attention (GQA) reduces memory demands by sharing attention weights across queries, thereby improving computational efficiency without sacrificing performance \cite{ainslie2023gqa}. Similarly, Flash Attention restructures attention computation by eliminating redundant operations, leading to measurable gains in both training speed and energy consumption \cite{golden2024flash}. These mechanisms have been successfully deployed in domains such as real-time translation, scientific research, and enterprise automation, demonstrating their practical efficacy in resource-constrained settings \cite{parthasarathy2024ultimate}. They have also been integrated into various open-source LLMs, enabling significant reductions in memory and energy consumption while maintaining high levels of accuracy and adaptability.

As such, on the question of innovations in development, the open-source approach emerges as the superior approach. While closed-source LLMs have historically driven foundational breakthroughs through large-scale resource investment, open-source LLMs have demonstrated that architectural ingenuity, efficiency-driven methodologies, and democratized participation are not only viable but preferable for long-term ML sustainability. By prioritizing modular development, parameter-efficient fine-tuning, and collaborative innovation, open-source frameworks ensure that technological advancements benefit a broader research ecosystem.

\section{Performance}
Evaluating LLMs requires a structured approach that examines their syntactic fluency, semantic coherence, and pragmatic efficacy across a wide range of linguistic and computational applications. Performance metrics are typically anchored in five primary dimensions: linguistic proficiency, inferential reasoning, domain-specific adaptability, computational efficiency, and resilience to adversarial perturbations and systemic biases \cite{mizrahi2024state}. Notably, standardized benchmarking protocols, including HumanEval, GSM8K, MMLU, DROP, and GPQA-Diamond, provide robust empirical frameworks for a comparative analysis of different model architectures.

Closed-source LLMs have historically exhibited superior performance on these benchmarks, largely attributable to their access to expansive proprietary datasets and sophisticated optimization paradigms. Models such as GPT-4 leverage pre-training corpora comprising hundreds of terabytes of textual data \cite{wang2023openchat}, thereby facilitating broad generalization with minimal task-specific fine-tuning \cite{ahmed2024studying}. The architectural sophistication of such models, typified by the incorporation of a trillion parameters and augmented by advanced inferential heuristics such as chain-of-thought prompting, facilitates exceptional proficiency in multi-step reasoning and complex problem-solving tasks \cite{wu2024comparative}. Moreover, Anthropic’s Claude prioritizes ethical alignment and safety, whereas Google’s Gemini integrates multimodal processing to achieve greater contextual awareness. However, these models are not without limitations. Data contamination—wherein overlaps between training and evaluation datasets artificially inflate benchmark metrics—raises concerns regarding artificially inflated performance metrics \cite{balloccu2024leak}, undermining the integrity of benchmark comparisons. Furthermore, the opacity of proprietary datasets and training methodologies precludes independent scrutiny, restricting transparency and impeding the replicability of claimed advancements \cite{liesenfeld2023opening}.

By contrast, open-source LLMs have made significant strides in closing the performance gap, particularly through advancements in computational efficiency and domain-specific fine-tuning. As such, LLaMA-2 achieves high linguistic fluency while reducing memory overhead through grouped query attention. Likewise, Mistral-7B outperforms GPT-3.5 across multiple linguistic benchmarks despite its comparatively smaller parameter scale. BLOOM extends multilingual capabilities, while specialized models such as BioBERT and LegalBERT exhibit superior performance in biomedical and legal text processing, respectively. It is particularly salient that DeepSeek-V3’s Mixture-of-Experts (MoE) framework selectively activates 37 billion of its 671 billion parameters per forward pass, optimizing computational efficiency without compromising performance. This architectural refinement has yielded competitive results on MMLU (88.5\% exact match), DROP (91.6\% F1 score), and GPQA-Diamond (59.1\% pass rate) \cite{rein2023gpqa}. Its mathematical and coding capabilities are equally notable, achieving a 90.2\% score on MATH-500 and an 82.6\% pass rate on HumanEval-Mul.

From Figure 1, it is evident that all models perform similarly on the MMLU metric, demonstrating that open-source models have significantly narrowed the performance gap with closed-source counterparts. Notably, Claude 3.5 Sonnet and DeepSeek-V3 exhibit higher performance on GSM8K compared to other models, highlighting their strength in mathematical reasoning tasks \cite{cobbe2021training}. Meanwhile, GPT-4’s consistent performance across all metrics underscores the benefits of expansive datasets and sophisticated architectures. However, the standout performance of LLaMA 3.1 405B on DROP suggests that open-source models are excelling in specific domains, leveraging optimized parameter efficiency \cite{dua2019drop}.

Further, techniques such as Low-Rank Adaptation (LoRA) and Conditioned Reinforcement Learning Fine-Tuning (C-RLFT) have been instrumental in LLMs’ domain-specific capabilities. LoRA’s ability to selectively fine-tune parameters has resulted in competitive results on benchmarks such as GSM8K \cite{hong2022analysis}. Likewise, NVIDIA’s NVLM 1.0D 72B model achieved a 4.3-point improvement in mathematical reasoning and coding tasks through multimodal training \cite{dai2024nvlm}. Unlike models such as InternVL2-Llama3-76B, which exhibit degraded text-based performance after multimodal training, NVLM not only preserves but also enhances its textual capabilities. This robustness enables NVLM to handle complex domain-specific inputs, such as handwritten pseudocode or location-specific queries, underscoring its precision within specialized contexts.

Despite these advancements, open-source LLMs face persistent challenges, particularly in benchmarking and resource allocation. The narrower scope of available training datasets and the limited computational resources accessible to the open-source community create systemic disparities in model evaluation \cite{lee2021landscape}. While closed-source models continue to dominate general-purpose NLP performance due to their access to proprietary data and extensive computational resources, open-source models are increasingly narrowing this gap through architectural ingenuity (via parameter-efficient optimization strategies) and collaborative research (via community-driven resource pooling) \cite{han2024parameter}. For these reasons, the open-source approach has the potential to achieve performance parity—or even surpass closed-source alternatives—insofar as it remains contingent upon sustained financial investment, access to high-quality training corpora, continuous advancements in optimization methodologies, as well as the development of parameter-normalized and task-agnostic evaluation frameworks that facilitate more equitable and comprehensive performance comparisons.

\begin{figure}[ht]
    \centering
    \includegraphics[width=1\linewidth]{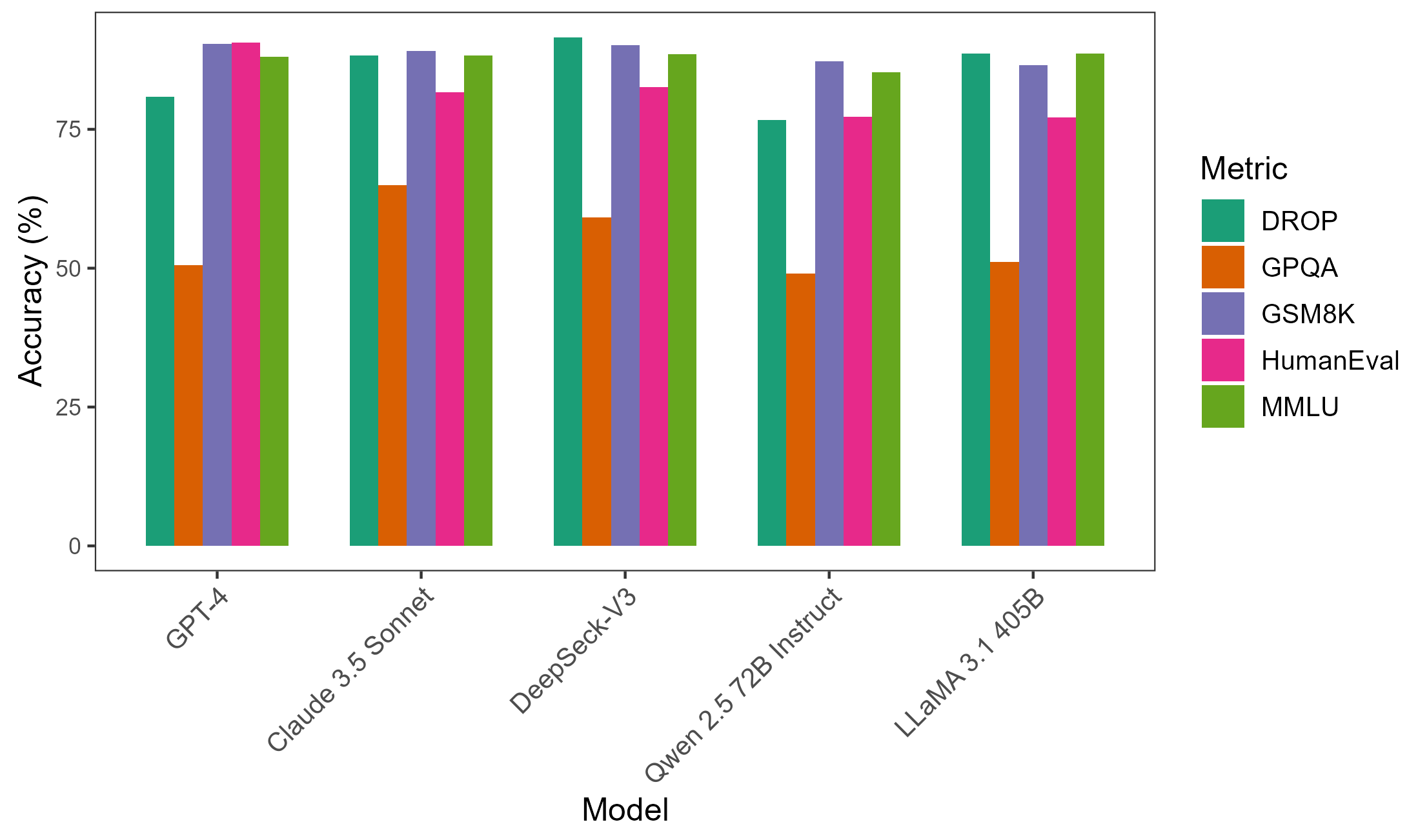}
    \caption{Benchmark performance of major LLMs on MMLU, GSM8K, HumanEval, DROP, and GPQA-Diamond as of October 2025.}
    \label{fig:plot}
\end{figure}

\section{Reproducibility}
Reproducibility is a fundamental metric for evaluating the reliability and validity of LLMs \cite{bhaskar2024reproscreener}. It refers to the ability of independent researchers to replicate the results of a model using the same data, code, and training conditions. In scientific research and engineering, reproducibility is a cornerstone of credibility, allowing for independent verification of findings, iterative improvements, and broader applicability of methodologies. 

By their very nature, closed-source LLMs pose marked challenges to reproducibility. The proprietary nature of these models means that external researchers cannot access hyperparameter configurations or optimization strategies. A case in point is OpenAI’s decision to withhold details about GPT-4’s architecture, pre-training corpora, dataset composition, and parameter count. Fine-tuning techniques like Reinforcement Learning from Human Feedback (RLHF) introduce another layer of opacity, as their efficacy is difficult to evaluate without access to the raw feedback data and optimization criteria. In so doing, they limit rigorous third-party evaluation and make it infeasible to assess whether performance gains stem from architectural innovations or merely from scaling up training data and compute resources. These practices have been criticized for the artificial inflation of benchmark performance via hidden biases, data contamination, and synthetic data augmentation. 

However, closed-source models can still contribute to reproducibility in specific domains that prioritize controlled, task-specific outputs. For example, Codex, which powers GitHub Copilot, has automated code generation \cite{chen2021evaluating}. and now allows developers to quickly prototype and refine software solutions. It has since lowered the barrier for non-experts while enhancing productivity for experienced programmers. So, although the model remains closed-source, its ability to generate consistent and predictable outputs for well-defined programming tasks highlights a form of reproducibility within constrained use cases. 

By contrast, open-source LLMs prioritize reproducibility by making their architectures, training data, and fine-tuning methodologies publicly available. This transparency enables independent verification, facilitates cross-institutional collaboration, and fosters continuous improvement. Models such as Meta’s LLaMA-2 and EleutherAI’s GPT-NeoX exemplify this commitment to reproducibility. LLaMA-2, in particular, has been reproduced by multiple research institutions and independent developers. They have utilized openly shared checkpoints and fine-tuning frameworks to verify its capabilities and adapt it to specialized applications. One prominent example is Hugging Face’s BLOOM, which was developed with an entirely transparent pipeline \cite{workshop2022bloom}. and has been successfully reproduced and fine-tuned by thousands of researchers worldwide, leveraging its open access to training data and model weights. The model’s training dataset, The Pile, is publicly accessible, allowing researchers to scrutinize data quality, distributional biases, and pre-processing steps \cite{gao2020pile}. Likewise, the training logs and hyperparameter configurations are openly available, making it possible to replicate training procedures on independent hardware setups. This level of transparency has enabled external research teams to fine-tune and extend BLOOM for domain-specific applications, validating its generalizability across diverse NLP tasks. BLOOM also supports 46 natural languages and 13 programming languages, making it one of the most globally inclusive open-source models \cite{le2023bloom}. 

Another approach to enhancing reproducibility in open-source LLMs is the use of standardized benchmarks and dataset curation practices. The BigScience project employed rigorous documentation protocols, ensuring that model performance is assessed under well-defined conditions \cite{daull2023complex}. The use of openly available evaluation suites, such as HELM (Holistic Evaluation of Language Models), further reinforces reproducibility by providing a common framework for comparing model outputs across different architectures and training paradigms \cite{liang2022holistic}. Through these mechanisms whose underlying ethos prioritizes collaboration, open-source LLMs  surpass their closed-source counterparts on the question of reproducibility.

\section{Transparency}
Transparency is a critical metric for evaluating the development, deployment, and governance of LLMs \cite{liao2023ai}. It refers to the degree to which a model’s internal mechanisms are accessible, interpretable, and verifiable by external researchers, policymakers, and end users. Transparency, therefore, is not merely a technical consideration but a moral obligation in the deployment of ML-driven decision-making systems. This metric rests on several key assumptions. First, it assumes that ML systems, particularly LLMs, can and should be made interpretable without compromising their functionality \cite{singh2024rethinking}. Second, it assumes that increased visibility into a model’s inner workings enables better scrutiny of biases, errors, and unintended consequences \cite{nabben2024ai}. Third, it assumes a tradeoff between proprietary protection and open scrutiny—while full disclosure may not always be feasible due to intellectual property concerns, selective transparency (e.g., model cards, algorithmic audits, and external oversight) can still enhance accountability \cite{mokander2024auditing}. The inputs to evaluating transparency include the availability of training data documentation, model architecture specifications, hyperparameter settings, and fine-tuning methodologies. The output of this metric manifests in tangible indicators such as open-source accessibility, comprehensive model documentation, the publication of research papers detailing methodologies, and third-party reproducibility. In balancing transparency across inputs and outputs, high transparency ensures that ML systems are less susceptible to biases, perform reliably across diverse contexts, and maintain accountability to stakeholders.

Closed-source LLMs, at least sometimes and in some cases, withhold critical information regarding model architectures, hyperparameter configurations, training datasets, and fine-tuning methodologies. The primary justification for restricted transparency in proprietary models is that disclosing training corpora and optimization strategies could render models vulnerable to adversarial attacks, facilitate model theft, or expose confidential proprietary methodologies \cite{kilic2024into}. However, the absence of disclosure requirements in proprietary models also means that developers are not obligated to detail the risks and mitigation strategies associated with their systems, limiting external accountability. This restricted access is particularly concerning in high-stakes domains involving critical infrastructure such as healthcare and legal decision-making, where algorithmic biases and errors can have profound consequences \cite{haltaufderheide2024ethics}. For example, when closed-source models produce outputs that reinforce harmful stereotypes, as was the case with Google’s Gemini model, it remains unclear whether the issue stems from biased training data, flawed objective functions, or other systemic deficiencies \cite{gautam2024melting}. The inability to scrutinize these aspects leaves stakeholders without recourse to challenge harmful biases, exacerbating trust deficits in ML deployment.

Conversely, open-source LLMs embrace transparency as a foundational principle. These models provide full access to model architectures, training data, and fine-tuning methodologies, allowing independent researchers to scrutinize, replicate, and refine them \cite{cai2024demystifying}. This approach facilitates rigorous external audits, enhances the interpretability of results, and democratizes ML research by enabling participation from a diverse array of contributors. The transparency inherent in open-source models confers significant advantages, particularly in the domains of bias mitigation and collaborative innovation. By making training datasets and decision-making processes accessible, open-source initiatives enable the systematic identification and rectification of biases that might otherwise remain undetected \cite{huang2023bias}. Their ethos also fosters a collaborative research ecosystem, wherein global researchers contribute to advancements in model efficiency, domain-specific applications, and ethical oversight.

However, the very openness of these models presents challenges; in particular, those concerning security vulnerabilities, external censorship, and misuse. Publicly accessible models are susceptible to adversarial exploitation, misinformation generation, and unintended biases introduced through poorly curated datasets. The case of DeepSeek’s R1 model exemplifies these risks, as evaluations revealed deliberate omissions of politically sensitive topics—such as the Tiananmen Square protests and Taiwan’s sovereignty—raising concerns about selective filtering and ideological bias in ostensibly transparent systems \cite{normile2025chinese}. Such redactions not only reflect political suppression but also introduce epistemic fragility, wherein ML models deprived of critical counterexamples and historical context fail to develop robust reasoning. As AI-mediated discourse increasingly shapes public narratives, the risk of censorship reinforcing state-controlled propaganda rather than facilitating open-ended inquiry becomes ever more pressing \cite{kronlund2024propaganda}.

These risks underscore the necessity for ethical oversight over LLM development. Mechanisms such as differential data access, responsible licensing agreements, and decentralized governance structures offer potential solutions to mitigate these concerns while preserving the benefits of transparency. In leveraging these strengths, on the question of transparency, open-source LLMs exhibit a decisive advantage over their closed-source counterparts.

\subsection{Community Contributions}

Community contributions are a cornerstone in the ongoing advancement, refinement, and deployment of LLMs, shaping both their technical evolution and governance structures \cite{zhou2023examining}. These contributions span a diverse spectrum of engagement, encompassing direct architectural modifications, dataset curation, adversarial testing, and ethical oversight. The contributors themselves emerge from a broad cross-section of expertise, including independent developers, academic researchers, industry professionals, and volunteers \cite{cheng2019activity}. Their motivations are equally varied—ranging from the pursuit of scientific inquiry and technological advancement to career development, intellectual curiosity, and open-source advocacy. Crucially, these contributions manifest at different junctures in the LLM lifecycle, from initial experimentation and architectural innovation to iterative fine-tuning, real-world application, and ongoing evaluation.

In closed-source ecosystems, participation is inherently constrained by proprietary boundaries that restrict external interventions to peripheral activities. The primary mechanisms for external engagement are interfaces such as APIs, fine-tuning frameworks, and plugin-based extensions, which allow third parties to adapt pre-trained models to domain-specific applications without altering their fundamental structure \cite{huang2023let}. For instance, OpenAI’s ChatGPT ecosystem supports limited customization through fine-tuning and prompt engineering, enabling users to modulate outputs within predefined constraints. However, such interactions remain extrinsic to the model’s core architecture, limiting the scope of community-driven enhancements. Moreover, while security researchers and ethical auditors endeavor to assess proprietary models for bias and vulnerabilities, their ability to effect meaningful change is curtailed by the opacity of these closed systems. Although centralized control over proprietary LLMs ensures stability and consistent performance, it simultaneously stifles opportunities for collaborative innovation and external scrutiny, inhibiting a collective approach to ML progress.

By contrast, open-source LLMs operate within a fundamentally different paradigm—one that embraces decentralized participation, modularity, and transparency. This ethos enables broad-based contributions to model architecture, training methodologies, and optimization strategies. A paradigmatic instance of this collective effort is the Aya model, a massively multilingual LLM that supports 101 languages, many of which are historically underrepresented in computational linguistics \cite{ustun2024aya}. This achievement underscores the capacity of open collaboration to expand model inclusivity and mitigate both linguistic and cultural disparities in ML development often overlooked by proprietary systems.

In addition to model and dataset enhancements, open-source ecosystems foster rigorous adversarial testing, ensuring robustness against manipulative or edge-case inputs. Researchers systematically stress-test these models to identify and rectify vulnerabilities, strengthening their adaptability to a variety of operational conditions. For instance, adversarial evaluation of hate speech detection systems has led to iterative refinements that enhance resilience against linguistic obfuscation techniques. Concurrently, open-source platforms function as repositories for knowledge dissemination, with initiatives such as Hugging Face and GitHub providing extensive documentation, pre-trained weights, and interactive tutorials \cite{pepe2024hugging}. These educational resources lower barriers to entry, democratizing access to state-of-the-art ML and fostering interdisciplinary engagement, particularly for individuals in resource-constrained environments. By engaging contributors across all skill levels—from novices to seasoned experts—these initiatives facilitate participation in dataset annotation, fine-tuning, and bug reporting, thereby fostering a collaborative and inclusive research ecosystem.

Despite their many advantages, open-source contributions introduce unique challenges, particularly in ensuring the sustainability and quality of community-driven efforts. The open-access nature of these models gives rise to the free-rider problem, where a significant portion of users benefit from shared resources without reciprocating contributions \cite{baldwin2003architecture}. This dynamic places undue strain on core maintainers, who bear the responsibility for repository management, quality control, and security oversight. To address these concerns, initiatives such as AI Commons have emerged, establishing collaborative infrastructure-sharing agreements that balance openness with long-term sustainability. Moreover, strategic partnerships with academic institutions, non-profits, and industry stakeholders bolster the resilience of open-source LLMs, ensuring the longevity and impact of collective innovation. These collaborations enhance scalability and inclusivity, preventing the burden from falling disproportionately on a small subset of contributors while ensuring that open-source models remain viable and impactful in the long term.

\section{Alternative Views}
The divide between open-source and closed-source approach to LLM development remains an unresolved and evolving discourse, with no immediate resolution favoring one paradigm unequivocally. In this section, we examine two alternative viewpoints: the sustained dominance of closed-source models due to their structural advantages and the potential emergence of hybrid models as a reconciliation of these opposing paradigms.

Despite sustained advocacy for open-source LLMs, the empirical and economic advantages afforded by closed-source models ensure their continued prevalence. These models have access to substantial computational resources and exclusive datasets, providing a pronounced competitive edge that will endure simply in virtue of their infrastructure. Their superiority is evident in standardized benchmarks such as MMLU, HumanEval, and GSM8K. Their exposure to high-quality and domain-specific pre-training corpora enhances generalization, particularly in complex reasoning and contextual adaptation. Furthermore, proprietary AI development benefits from robust financial investment. This funding sustains research into optimization methodologies that remain largely inaccessible to the open-source ecosystem.

Risk mitigation presents another compelling reason for the persistence of closed-source LLMs. Organizations operating in high-stakes domains—including healthcare, finance, and regulatory governance—prioritize operational reliability, security, and compliance adherence. Proprietary frameworks provide a controlled infrastructure that mitigates adversarial vulnerabilities. They also enable enterprises to enforce stringent oversight mechanisms. Moreover, integrating closed-source solutions within existing corporate workflows is streamlined by dedicated customer support and iterative performance enhancements. These aspects are inherently more challenging to sustain within decentralized open-source initiatives, and serve to justify the widespread enterprise adoption of closed-source models.

Nevertheless, increasing concerns regarding ethical AI governance and regulatory compliance are exerting pressure on proprietary LLM developers to introduce elements of transparency. As scrutiny over ML accountability intensifies, proprietary developers may be compelled to adopt partial disclosure measures. This shift could pave the way for hybrid models that incorporate aspects of both closed- and open-source approaches.

Hybrid models offer a synthesis by integrating the advantages of proprietary scalability with selective transparency. Under such frameworks, companies could disclose anonymized training methodologies, high-level abstractions of decision-making logic, or curated datasets for third-party evaluation \cite{von2024systematic}. Modular transparency mechanisms would enable independent audits while safeguarding corporations’ intellectual property. This approach strikes a balance between proprietary control and external accountability.

A salient example of a hybrid implementation is the increasing reliance on Application Programming Interfaces (APIs) that facilitate controlled access to proprietary LLMs. OpenAI’s GPT models, for instance, provide API-based integration pathways. These allow developers to leverage cutting-edge ML capabilities while maintaining safeguards against misuse. Furthermore, some enterprises have begun open-sourcing selective fine-tuning frameworks while preserving the opacity of their foundational architectures. This bifurcated approach allows researchers to experiment with adaptable model architectures without exposing core proprietary assets. It fosters a degree of collaborative innovation within a restricted ecosystem.

\section{Discussion}

The rise of open-source LLMs marks an important milestone in NLP  and has undoubtedly intensified the ongoing debate over the direction of ML research. As discussed earlier, closed-source models are often defended for their proprietary safeguards, security measures, and regulatory compliance, whereas open-source alternatives prioritize transparency, collaborative development, and accessibility. In this position paper, we have argued that open-source LLMs not only enhance transparency and reproducibility but also cultivate a more accessible, collaborative, and equitable research ecosystem. In fact, they contribute to the competitive landscape with iterative development that fosters rapid innovation in architectural efficiency, fine-tuning methodologies, and domain-specific applications. As such, open-source models demonstrate that decentralized ML research can rival—and, in some cases, surpass—the performance capabilities of their proprietary counterparts. This emerging dominance, of course, demands a rigorous discussion of the structural, economic, and ethical dimensions. In particular, while these models lower entry barriers for research and development, they simultaneously raise pressing concerns regarding their sustainability, vulnerability to misuse, and capacity to compete with proprietary solutions. Thus, it remains indeterminate whether open-source LLMs can remain sustainable and competitive in the long run as viable alternatives to closed-source models.

One of the foremost considerations in this discourse is the financial sustainability of open-source LLMs, especially given the economic advantages leveraged by closed-source models through proprietary monetization frameworks and corporate backing. This contrast underscores the challenge of ensuring continued funding for decentralized ML research. The operation and continuous refinement of such models require vast computational resources, posing substantial challenges for independent researchers and institutions with limited funding. In contrast to corporate-backed closed-source models, which leverage proprietary monetization frameworks to subsidize operational costs, open-source models rely on distributed funding mechanisms, including academic grants, philanthropic contributions, and collaborative resource-sharing initiatives. Future research must explore scalable economic frameworks, such as federated computing infrastructures and decentralized funding pools, to ensure the financial viability of open-source LLMs without undermining their accessibility and openness.

Beyond economic feasibility, open-source LLMs necessitate a careful evaluation of their implications for information integrity. The unrestricted availability of these models introduces a double-edged challenge: while they enable transparency and foster independent innovation, they also expose ML systems to adversarial manipulation, disinformation campaigns, and algorithmic biases. Unlike closed-source models, which implement centralized moderation mechanisms to mitigate these risks, open-source platforms must develop decentralized governance structures capable of safeguarding model integrity. Robust adversarial testing, standardized ethical auditing protocols, and interpretability mechanisms will be indispensable for ensuring that open-source LLMs maintain reliability while mitigating their susceptibility to propaganda and censorship concerns.

Moreover, open-source LLMs hold the potential to dismantle the AI oligopoly by decentralizing access to state-of-the-art models \cite{verkama1992multi}. However, their capacity to serve as viable alternatives to proprietary systems is contingent upon their ability to achieve comparable levels of performance, scalability, and enterprise adoption. Given the disparities in computational resources between open-source initiatives and corporate entities, sustaining competitive innovation in open-source LLMs necessitates novel efficiency-enhancing techniques. Research in parameter-efficient fine-tuning, energy-efficient training paradigms, and scalable distributed inference strategies will be integral to ensuring that open-source models remain both high-performing and widely deployable. It is critical to note that the rate at which open-source models evolve is directly proportional to the level of community contribution they receive. Unlike closed models with centralized control, open-source initiatives thrive on collective effort, where advancements in both architecture and training rely on the sustained engagement of researchers, developers, and institutions. This symbiotic relationship underscores the necessity of fostering a robust ecosystem of contributors to sustain long-term progress.

The trajectory of NLP development is now at an inflection point on the broader normative question of whether to embrace open-source or closed-source approaches. While open-source LLMs hold promise for democratizing technological progress, their sustainability hinges on resolving governance, funding, and performance-related challenges. However, their long-term success is contingent on financial sustainability, governance mechanisms, and competitive performance. Future research must prioritize strategic interventions that preserve the integrity and scalability of open-source models, ensuring their role as a formidable and ethically grounded alternative to proprietary models. The discourse must therefore evolve beyond the binary of open- versus closed-source to examine instead how open-source LLMs can be designed to endure as sustainable, responsible, and robust models of the future of machine learning.

\section*{Acknowledgments}
This work was supported, in part, by contributions from the open-source AI community.

\bibliographystyle{unsrt}  
\bibliography{references}  

\end{document}